\documentclass[10pt,twocolumn,letterpaper]{article}

\usepackage{cvpr}              % To produce the CAMERA-READY version
\usepackage[dvipsnames]{xcolor}

\definecolor{cvprblue}{rgb}{0.21,0.49,0.74}
\usepackage[pagebackref,breaklinks,colorlinks,citecolor=cvprblue]{hyperref}
\usepackage[accsupp]{axessibility}

\def\paperID{12} % *** Enter the Paper ID here
\def\confName{CVPR}
\def\confYear{2024}

\title{Refining Remote Photoplethysmography Architectures using CKA and Empirical Methods}

\author{Nathan Vance and Patrick Flynn\\
University of Notre Dame\\
{\tt\small \{nvance1,flynn\}@nd.edu}
}

\begin{document}

\maketitle

\begin{abstract}
Model architecture refinement is a challenging task in deep learning research fields such as remote photoplethysmography (rPPG). One architectural consideration, the depth of the model, can have significant consequences on the resulting performance. In rPPG models that are overprovisioned with more layers than necessary, redundancies exist, the removal of which can result in faster training and reduced computational load at inference time. With too few layers the models may exhibit sub-optimal error rates. We apply Centered Kernel Alignment (CKA) to an array of rPPG architectures of differing depths, demonstrating that shallower models do not learn the same representations as deeper models, and that after a certain depth, redundant layers are added without significantly increased functionality. An empirical study confirms how the architectural deficiencies discovered using CKA impact performance, and we show how CKA as a diagnostic can be used to refine rPPG architectures.
\end{abstract}
\section{Introduction}

Remote Photoplethysmography (rPPG) is a technique for inferring the pulse waveform of a subject using %a video camera.
digital video sequences of the subject's skin (usually the face). Recent advances in rPPG have employed deep 3D convolutional neural networks (3DCNNs) with great success, achieving error rates of less than 5 BPM in challenging scenarios~\cite{yu2019remote,speth2021unifying}.

It is important and informative to probe the architectures of rPPG models to understand their critical and redundant portions. If a model is too shallow it may underperform. However, a deep model with many redundant layers requires more computational resources to load and train than a shallower counterpart with redundancies stripped.

The fine-tuning of architecture depths may be achieved through a brute-force parameter sweep (i.e., train a wide array of architectures and select the one that minimizes errors). However, it is informative to understand why shallower and deeper models fail while others succeed, as this informs further architectural refinements.

In this study we make the following contributions:

\begin{itemize}
    \item We develop an array of PhysNet-3DCNN~\cite{yu2019remote} variants ranging from 2 to 15 layers in depth, as well as TS-CAN~\cite{liu2020multi} variants ranging from 1 to 10 meta-layers in depth.
    \item We perform a Centered Kernel Alignment (CKA)~\cite{kornblith2019similarity} analysis to yield insights into network pathologies, revealing both network redundancies and also critical representations.
    \item We demonstrate that the findings from the CKA analysis are reflected in empirical results, arguing that such techniques should be used by the research community to refine network architectures.
\end{itemize}

\begin{figure}
    \centering
    \includegraphics[width=\linewidth]{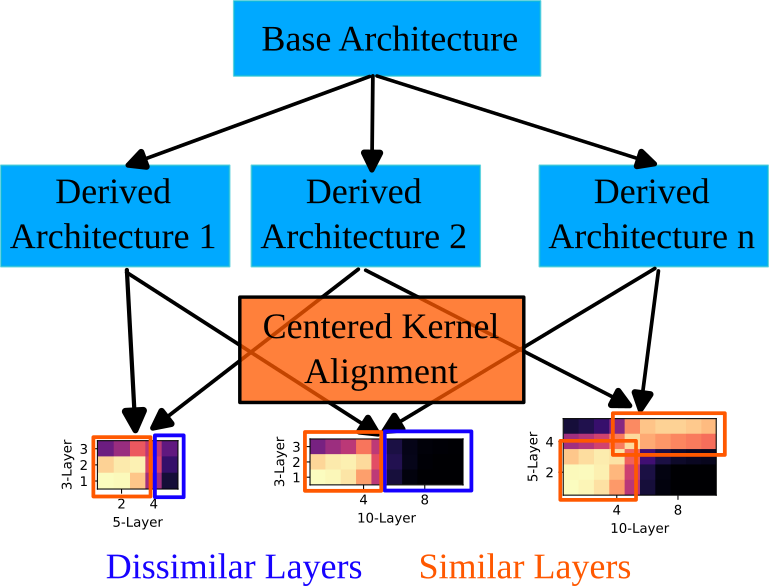}
    \caption{Overview of using CKA to inform architecture refinement by revealing similar and dissimilar layers between architectures.}
    \label{fig:overview}
\end{figure}

Figure \ref{fig:overview} shows an overview of our workflow. Using CKA, we compare architectures, determining which layers correspond to each other across architectures, and which contain representations unique to that architecture.
\section{Related Work}

\subsection{Remote Photoplethysmography}

Remote Photoplethysmography (rPPG) is a technique to infer a subject's pulse waveform from video data, pioneered by Takano and Ohta in \cite{takano2007heart} and Verkruysse et al. in \cite{verkruysse2008remote}. Poh et al. developed an early technique using blind source separation~\cite{poh2010non}. Noteworthy classical techniques include CHROM (developed by deHaan and Jeanne~\cite{DeHaan2013}), which was designed to be robust against motion, and POS (developed by Wang et al.~\cite{Wang2017}), which relaxes assumptions made in CHROM regarding skin tone.

Deep learning techniques for rPPG have proliferated due to their success at this task, and can be divided into two general camps: frame-difference processing architectures which compute the rPPG derivative between frames~\cite{Chen2018,liu2020multi,Zhao_2021,Niu2020,lu2021dual}, and sequence processing architectures which perform rPPG over an entire clip in an end-to-end fashion~\cite{yu2019remote,tsou2020siamese,Lee_ECCV_2020,speth2021unifying,Yu_2022_CVPR}.

Frame-difference processing architectures are among the earlier deep learning rPPG systems to be developed and have retained relevancy as lightweight alternatives to end-to-end techniques. Chen and McDuff developed DeepPhys, a 2DCNN-based architecture with two branches, one processing frame differences and the other processing raw frames~\cite{Chen2018}. Liu \etal expanded the capabilities of the DeepPhys architecture with TS-CAN which is capable of multi-task learning of physiological signals such as respiration and the pulse waveform~\cite{liu2020multi}. Zhao \etal also developed this dual branch architecture by employing a 3D central difference convolution operation for noise reduction~\cite{Zhao_2021}. Some additional non-end-to-end architectures utilize spatial-temporal maps, including the work of Niu \etal which passes these maps into a ResNet-18 backbone followed by a gated recurrent unit to predict the heart rate directly~\cite{Niu2020}, and Dual-GAN developed by Lu \etal which models both the pulse waveform and its noise distribution using two GAN modules~\cite{lu2021dual}. Due to the availability of an implementation of TS-CAN in the open source rPPG-Toolbox~\cite{liu2022rppg}, and its adoption as a state-of-the-art baseline for rPPG research, we select TS-CAN as the frame-difference architecture to which we apply our techniques.

In 2019 Yu \etal explored the use of end-to-end architectures in rPPG by developing PhysNet in two variants, an end-to-end variant based on 3DCNNs, and a non-end-to-end variant that uses 2DCNN feature extractors followed by a recurrent network~\cite{yu2019remote}. The 3DCNN structure was explored by Lee \etal with Meta-rPPG, which utilizes an hourglass encoder-decoder structure which forces the network to consider the entire timeframe~\cite{Lee_ECCV_2020}. Tsou \etal additionally developed the 3DCNN architecture by using Siamese 3DCNN networks to jointly predict rPPG signals over the forehead and cheek, then combining the result into a single pulse waveform~\cite{tsou2020siamese}. Speth \etal proposed RPNet which relaxes assumptions the 3DCNN PhysNet architecture makes regarding framerate by including temporal dilations~\cite{speth2021unifying}. In addition, video transformer networks have also been investigated for rPPG, with Yu \etal developing PhysFormer~\cite{Yu_2022_CVPR}. In this work we apply our techniques to PhysNet-3DCNN because it is an early work on which several end-to-end rPPG architectures are based.

\subsection{Neural Network Similarity}

The comparison of neural network representations is an important task for understanding their underlying pathology. Early pioneers in this area include Laakso and Cottrell, who base their network comparison on distance between network activations~\cite{laakso2000content}. Raghu et al. developed a technique called Singular Vector Canonical Correlation Analysis (SVCCA) that allows network comparisons between different layers and architectures~\cite{raghu2017svcca}. Morcos et al. build on SVCCA, proposing Projection Weighted CCA (PWCCA), which better differentiates between signal and noise~\cite{morcos2018insights}. Kornblith et al. propose using Centered Kernel Alignment (CKA) for network similarity analysis, which they find is more reliable in light of different network initializations relative to CCA-based approaches~\cite{kornblith2019similarity}. Other families of methods exist, most notably Representational Similarity Analysis (RSA), which is used heavily by the neuroscience research community~\cite{kriegeskorte2008representational}.

Cui \etal provide some critique on CKA and RSA, finding that they may indicate high similarity in random networks or perform inconsistently in transfer learning, and propose modifications to resolve these issues~\cite{cui2022deconfounded}. However, our analysis in Section \ref{sec:cka_method} verifies that CKA reveals the behavior of models as required from an effective diagnostic tool for architecture refinement.
\section{Methods}

\subsection{Flexible Depth Models} ~\label{sec:flex}

We performed a parameter sweep over the 3DCNN based PhysNet architecture developed by Yu et al.~\cite{yu2019remote}, and TS-CAN developed by Liu et al.~\cite{liu2020multi}, by varying the depth of the networks. In this section we discuss modifications made to the published architectures to facilitate this study.

\subsubsection{PhysNet-3DCNN}

PhysNet-3DCNN, or simply 3DCNN, is an rPPG architecture utilizing 10 Conv3d layers operating over video frame sequences. In particular, the input video is cropped around the face and downsampled to $64\times64$ pixels using cubic interpolation, then fed into the network in groups of $T$ consecutive frames. The network outputs a $T$-length pulse waveform. In this work, the parameter $T$ was selected to be 136 frames as this value typically captures at least one full heart cycle, yet fits on GPUs up to the largest depth network evaluated, and 3DCNN has been shown to exhibit relatively little performance variability for values of $T$ between 32 and 256 frames~\cite{yu2019remote}.

We generated variations of 3DCNN for network depths of 2 to 15 Conv3d layers. In particular, as with the default 3DCNN, the first Conv3d layer has a (1,5,5) kernel size and 32 output channels, all intermediate Conv3d layers are (5,3,3) with 64 output channels, and the last is (1,1,1) with 1 output channel. All Conv3d layers except for the final layer are followed by a batch normalization layer and a ReLU. Odd numbered layers other than the first layer are further followed by a drop3d layer configured at p=0.5.

In order to accommodate varying numbers of layers, we organize pooling layers after the ReLU or drop3d (if applicable) of certain conv3d layers as outlined in Table~\ref{tab:pooling-3dcnn}. In each case, the final pooling layer is an average pooling layer, while all others are maximum pooling layers.

\begin{table}
    \centering
    \begin{tabular}{c|cc}
        \toprule
        Depth & Pooling Indices & Spatial Stride \\
        \midrule
        2 & 1 & 64 \\
        3 & 1,2 & 8,8 \\
        4 & 1,2,3 & 4,4,4 \\
        5 & 1,2,3,4 & 2,4,2,4 \\
        6 & 1,2,3,4,5 & 2,2,2,2,4 \\
        7 & 1,2,3,4,5,6 & 2,2,2,2,2,2 \\
        8 & 1,2,3,4,5,7 & 2,2,2,2,2,2 \\
        9 & 1,2,3,4,6,8 & 2,2,2,2,2,2 \\
        10 & 1,2,3,5,7,9 & 2,2,2,2,2,2 \\
        11 & 1,2,4,6,8,10 & 2,2,2,2,2,2 \\
        12 & 1,3,5,7,9,11 & 2,2,2,2,2,2 \\
        13 & 2,4,5,8,10,12 & 2,2,2,2,2,2 \\
        14 & 3,5,7,9,11,13 & 2,2,2,2,2,2 \\
        15 & 4,6,8,10,12,14 & 2,2,2,2,2,2 \\
        \bottomrule
    \end{tabular}
    \caption{Pooling layer configuration for 3DCNN variants}
    \label{tab:pooling-3dcnn}
\end{table}

\subsubsection{TS-CAN}

The Temporal Shift Convolutional Attention Network (TS-CAN)~\cite{liu2020multi} was developed to jointly probe spatial and temporal features in video data. In particular, input video is cropped around the face and downsampled to $36\times36$ pixels, then the pairwise differences between frames are calculated. Both the raw frames and the pairwise differences are fed into the network in 20-frame segments. TS-CAN can be trained to produce a single rPPG sequence, or multiple sequences for multi-task learning. In our experiments we focus on the single-task rPPG problem.

We generated TS-CAN models of varying depths by grouping a set of operations into a ``meta-layer'', which we then repeat to depths of 1 to 10 meta-layers. The grouping is as follows:

\begin{itemize}
    \item Diff branch: TSM, Conv2d, tanh, TSM, Conv2d, tanh.
    \item Raw branch: Conv2d, tanh, Conv2d, tanh.
    \item Mixing branch: Conv2d (of raw branch output), sigmoid, attention mask, and results are multiplied by the diff branch output.
    \item Diff branch: average pooling (of mixing branch output), dropout.
    \item Raw branch: average pooling, dropout.
\end{itemize}

We configure layers in these groupings identically to the published TS-CAN model, an implementation of which is available in the rPPG-Toolbox~\cite{liu2022rppg}.

The TS-CAN architecture yields a decrease in spatial resolution at every meta-layer. As a result, in order to accommodate deeper variants we both increase the size of input video frames from the published $36x36$ pixels to $64x64$ pixels, and we constrain average pooling to select layers as given in Table~\ref{tab:pooling-tscan}.

\begin{table}
    \centering
    \begin{tabular}{c|c}
        \toprule
        Depth & Pooling Indices \\
        \midrule
        1 & 1 \\
        2 & 1,2 \\
        3 & 1,2,3 \\
        4 & 1,2,3,4 \\
        5 & 1,3,4,5 \\
        6 & 1,3,5,6 \\
        7 & 1,3 \\
        8 & 1,3 \\
        9 & 1,3 \\
        10 & 1 \\
        \bottomrule
    \end{tabular}
    \caption{Pooling layer configuration for TS-CAN variants}
    \label{tab:pooling-tscan}
\end{table}

\subsection{CKA Analysis} ~\label{sec:cka_method}

We performed a Centered Kernel Alignment (CKA) analysis for each model depth to understand network pathology. After~\cite{kornblith2019similarity}, we hypothesize that strong similarities between different layers of the same model indicate redundancies in the architecture such that the number of layers may be reduced without a large performance degradation. Furthermore, we hypothesize that highly similar layers between different architectures perform similar rPPG tasks, and that any layers without a corresponding similar layer in the other architecture may perform a task not handled by the other architecture. Examples of such groupings of similar and dissimilar layers are shown in Figure \ref{fig:overview}, and are expounded upon and analyzed in Section \ref{sec:cka-results}.

\begin{figure*}
    \centering
    \subfloat[3DCNN]{\includegraphics[width=.8\linewidth]{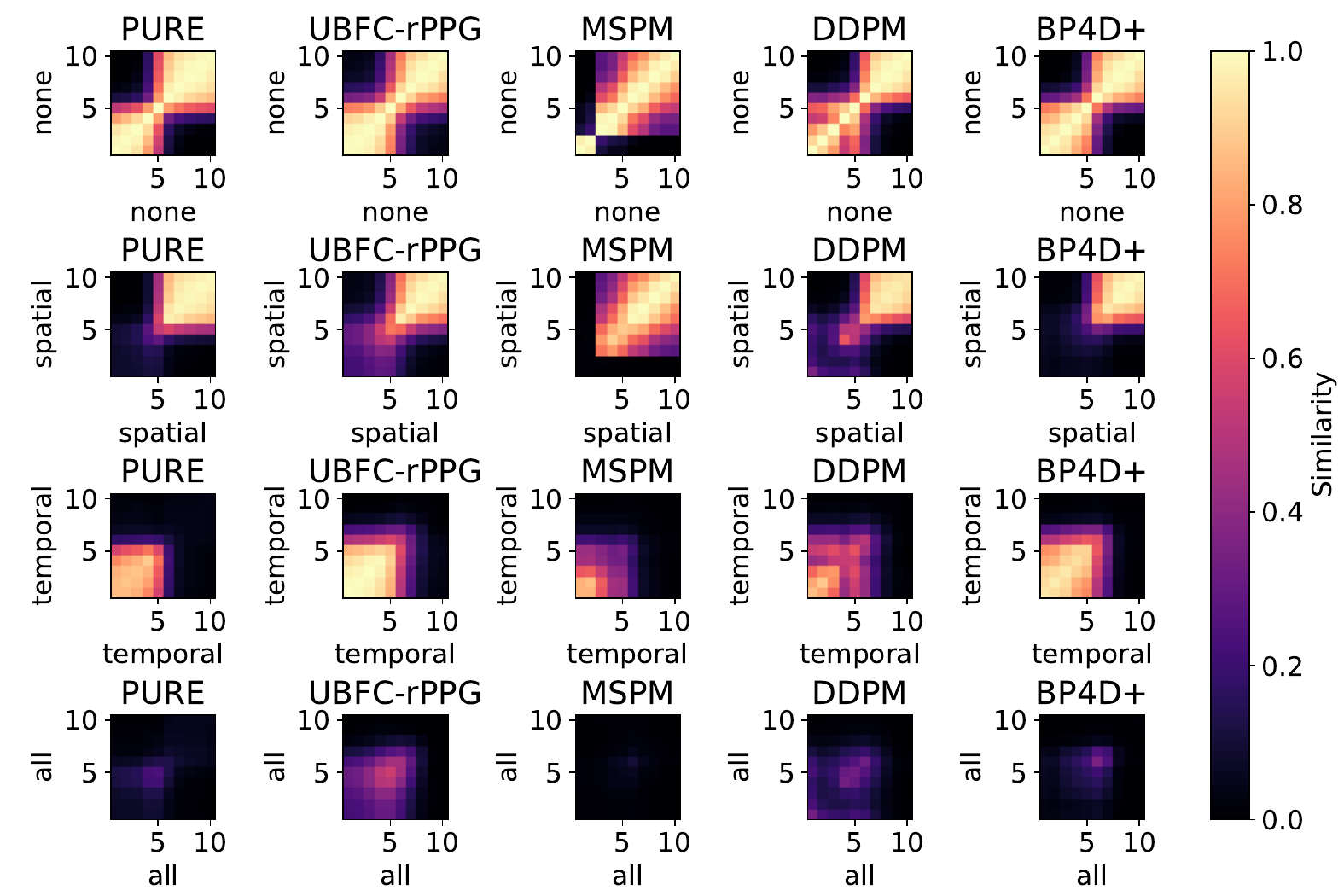}\label{fig:augs:3dcnn}} \\
    \subfloat[TS-CAN]{\includegraphics[width=.8\linewidth]{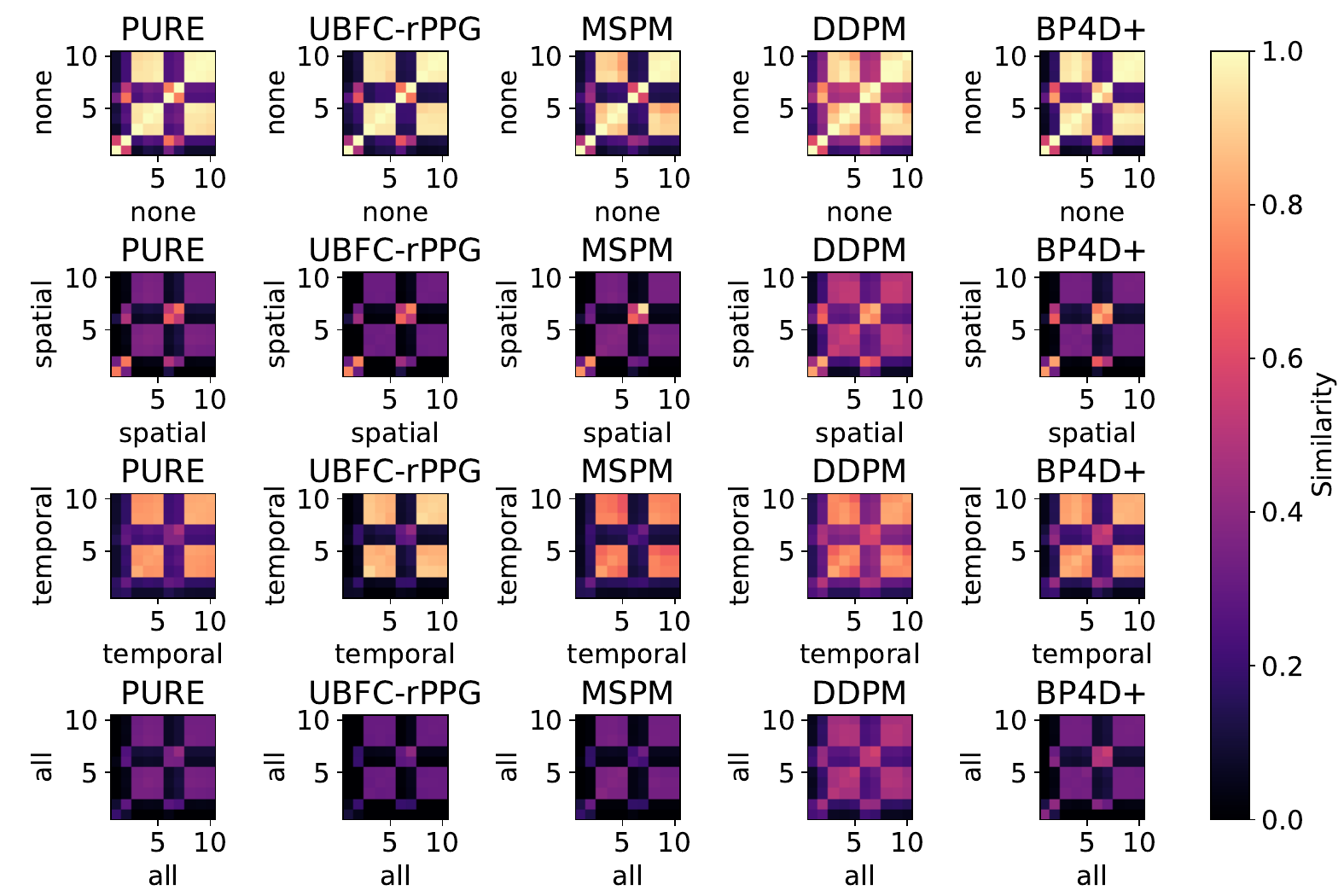}\label{fig:augs:tscan}}
    %\subfloat[PhysNet]{\includegraphics[width=\linewidth]{figs/cka-augs/physformer-all-cka.pdf}\label{fig:augs:physformer}} \\
    \caption{CKA comparison across augmentations, datasets, and architectures, demonstrating that blocks of layers exhibit similar behavior.}
    \label{fig:augs}
\end{figure*}

We provide an illustrative example of how CKA highlights model behavior. We trained each of the investigated architectures at their published depths (\ie, 3DCNN-10 and TS-CAN-2). The training was performed over a variety of datasets that are described in Section \ref{sec:results}, and we performed CKA analyses over them with different sets of transformations intended to showcase how portions of the network operate. In particular, this analysis comprises the following transformation sets:

\begin{itemize}
    \item \textbf{none}: Perform no transformations.
    \item \textbf{spatial}: Perform spatial transformations: randomly flip, add illumination noise, and Gaussian blur.
    \item \textbf{temporal}: Vary the playback speed and modulate the change in playback speed.
    \item \textbf{all}: Combine both spatial and temporal transformations.
\end{itemize}

Figure \ref{fig:augs} depicts CKA maps comparing the similarity of models. In particular, the similarity of activations of every layer in a model on the x axis is mapped against every layer of the model on the y axis. The numeric value on each axis corresponds to the layer index in the architecture as described in Section \ref{sec:flex}. Layer comparisons with a high degree of similarity result in a lighter color, whereas layer comparisons that are less similar result in a darker color.

Different portions of the network are affected differently by these transformations. In each column of Figure \ref{fig:augs}, a different dataset is explored, and in each row the model is compared to itself while undergoing one of four sets of transformations.

In Figure \ref{fig:augs:3dcnn} we observe that, across all datasets, the 3DCNN network adopts a block structure with two main regions which vary in size depending on the training dataset, as most clearly presented in the comparisons with no transformations (the first row of Figure \ref{fig:augs:3dcnn}). When the spatial transformations are applied, the similarity of the earlier region is reduced, while the latter region remains relatively unaffected. Similarly, the temporal transformations affect the latter region more heavily than the early region. When both transformation sets are applied the full model is affected. These results indicate that across datasets the 3DCNN model learns an internal structure in which its early layers process spatial features, while its latter layers process temporal features.

In Figure \ref{fig:augs:tscan} we observe a different internal structure in TS-CAN than in 3DCNN. This is due to the dual-branch architecture of TS-CAN, in which the first two layers are part of the ``Diff'' branch dealing with temporal features, then the next three layers are part of the ``Raw'' branch dealing with spatial features, then the pattern repeats for the second meta-layer.
%We observe that different layers perform distinct tasks, but unlike 3DCNN these are learned from the data, in TS-CAN each branch takes on the spatial or temporal task in keeping with its design.
This is observed in the spatial row of Figure \ref{fig:augs:tscan} in which the similarity within the ``Raw'' blocks is reduced while the ``Diff'' blocks are relatively less affected. Similarly the temporal augmentations heavily affect the ``Diff'' blocks while affecting the ``Raw'' blocks less severely.

These analyses indicate that both 3DCNN and TS-CAN have internal structures in which some layers process temporal features while others process spatial features. In the case of 3DCNN, the spatial processing layers occur early in the network, temporal layers occur later in the network, and the precise divide appears to be learned from the data. In the case of TS-CAN, the assignment of tasks is a facet of the dual-branch architecture. Furthermore, these analyses demonstrate how CKA reveals behavioral information regarding how models process data.
\section{Results} \label{sec:results}

\begin{figure*}
    \centering
    \subfloat[3DCNN]{\includegraphics[width=.49\linewidth]{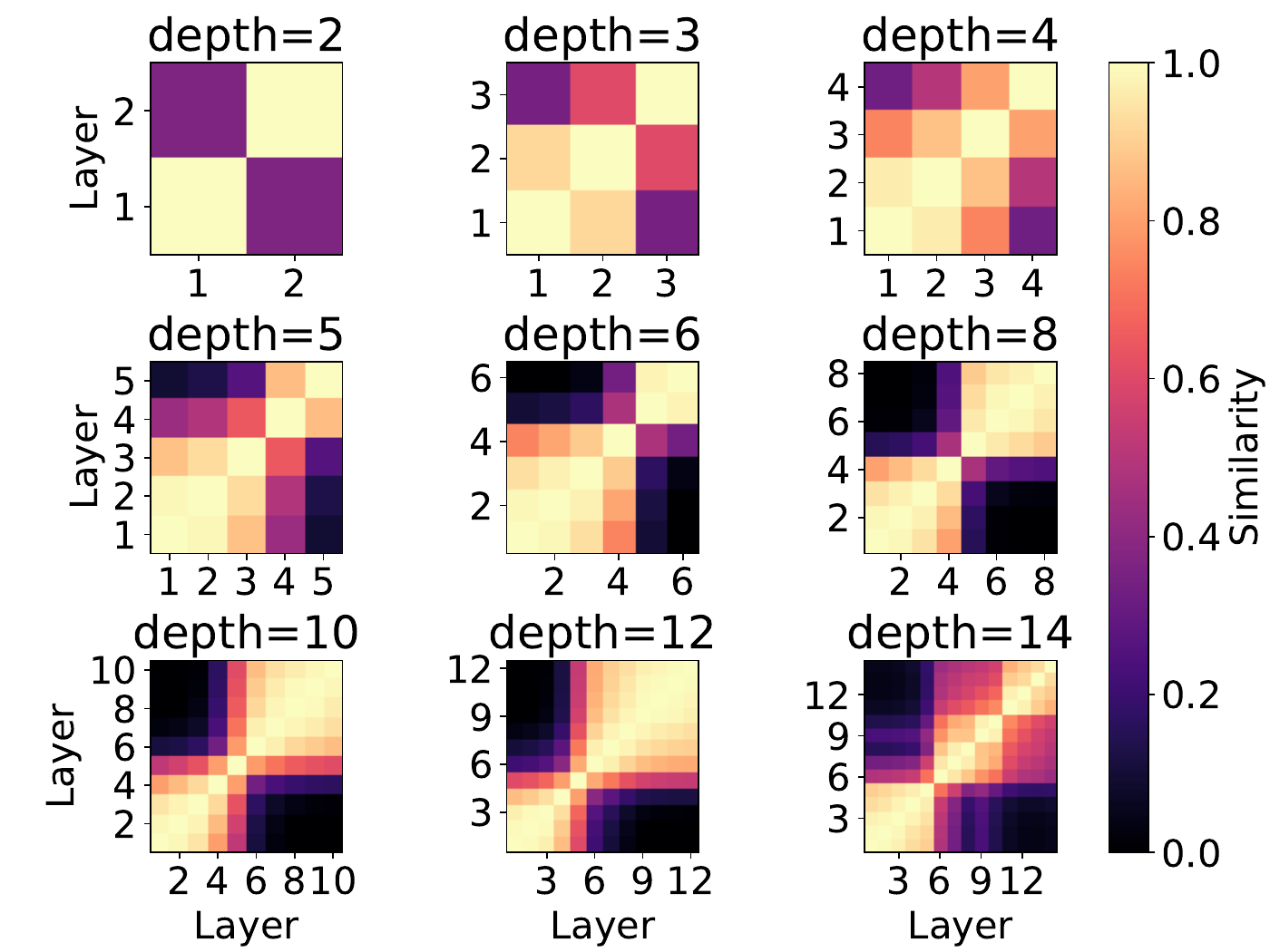}\label{fig:cka-self:3DCNN}}
    \subfloat[TS-CAN]{\includegraphics[width=.49\linewidth]{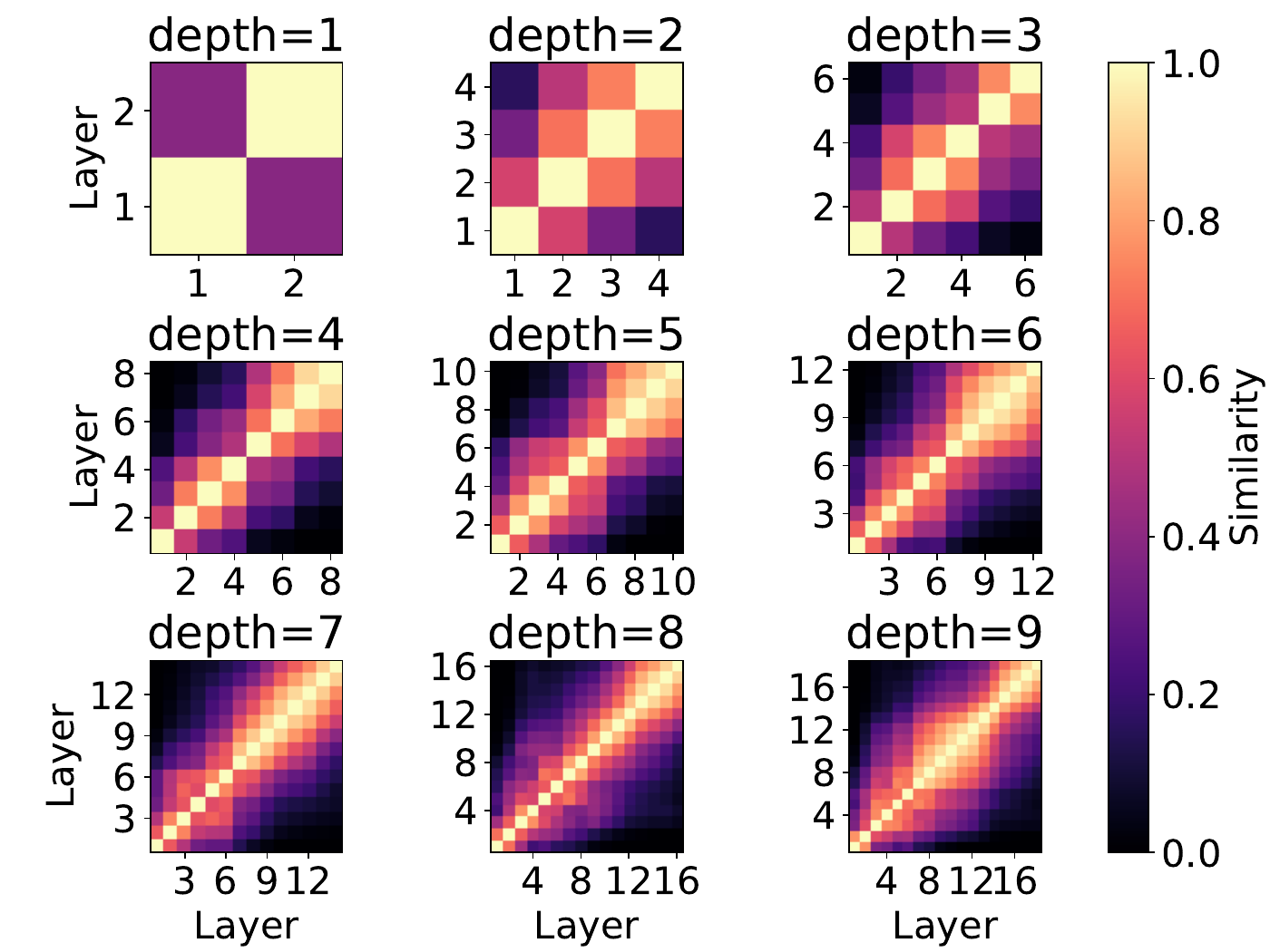}\label{fig:cka-self:TSCAN}}
    \caption{CKA self-similarity comparison for 3DCNN (\ref{fig:cka-self:3DCNN}) and TS-CAN (\ref{fig:cka-self:TSCAN}) based architectures on the PURE dataset.}
    \label{fig:cka-self}
\end{figure*}

We trained models based on 3DCNN and TS-CAN at different depths according to the architecture definitions given in Section~\ref{sec:flex}.  These models were then analyzed with CKA and also used in pulse estimation experiments to determine if architectural under/overprovisioning is reflected in empirical results. For a robust analysis, we investigated the following datasets:

\begin{itemize}
    \item \textbf{PURE}~\cite{Stricker2014} is a small dataset of 10 subjects with six one-minute videos, each with constrained head motions.
    \item \textbf{UBFC-rPPG}~\cite{Bobbia2019} is a 43 subject dataset with an average of 1.6 minutes of video for each subject. In each video, the subject played a time-sensitive mathematical game intended to elicit an elevated heart rate.
    \item \textbf{DDPM}~\cite{Vance2022} is a large 93 subject dataset with between 8 and 11 minutes of video for each subject. In each video, subjects engaged in a mock-interview in which they attempted to deceive the interviewer on selected questions.
    \item \textbf{MSPM}~\cite{niu2023full} is a large 103 subject dataset with an average of 14 minutes of video for each subject. In each video, subjects engaged in a sequence of activities including a breathing exercise, playing a racing game, and watching videos. The dataset includes an adversarial attack in which pulsating colored light is projected onto the subjects --- we omit this portion of the dataset in our analyses (the adversarial attack succeeds in obliterating the pulse waveform in its interval).
    \item \textbf{BP4D+}~\cite{zhang2016multimodal} is a large 140 subject dataset with an average of 9 minutes of video for each subject. This dataset is a spontaneous emotion corpus in which each subject experiences 10 different activities designed to elicit different emotions (\eg embarrassment due to improvising a silly song, or startle/surprise due to experiencing a sudden burst of sound). It is collected with a continuous blood pressure monitoring for its blood volume pulse ground truth, whereas the other datasets use a pulse oximeter.
\end{itemize}

We trained most models for 40 epochs. 3DCNN variants with depths of 14-15 being trained on DDPM and MSPM required 80 epochs for the loss to reach a plateau. We trained using the same set of augmentations as used in~\cite{vance2023promoting}, in which video clips are scaled temporally to capture a broad band of heart rate frequencies. This augmentation has been shown to promote generalization rather than memorization --- an important feature for meaningful network pathology analysis~\cite{morcos2018insights}. We use a negative Pearson loss function, k-fold cross validation with k=5, the Adam optimizer with a learning rate of 0.0001, and validation loss for model selection.

\subsection{CKA-based Analysis} \label{sec:cka-results}

\begin{figure*}
    \centering
    \subfloat[10-to-all]{\includegraphics[width=.5\linewidth]{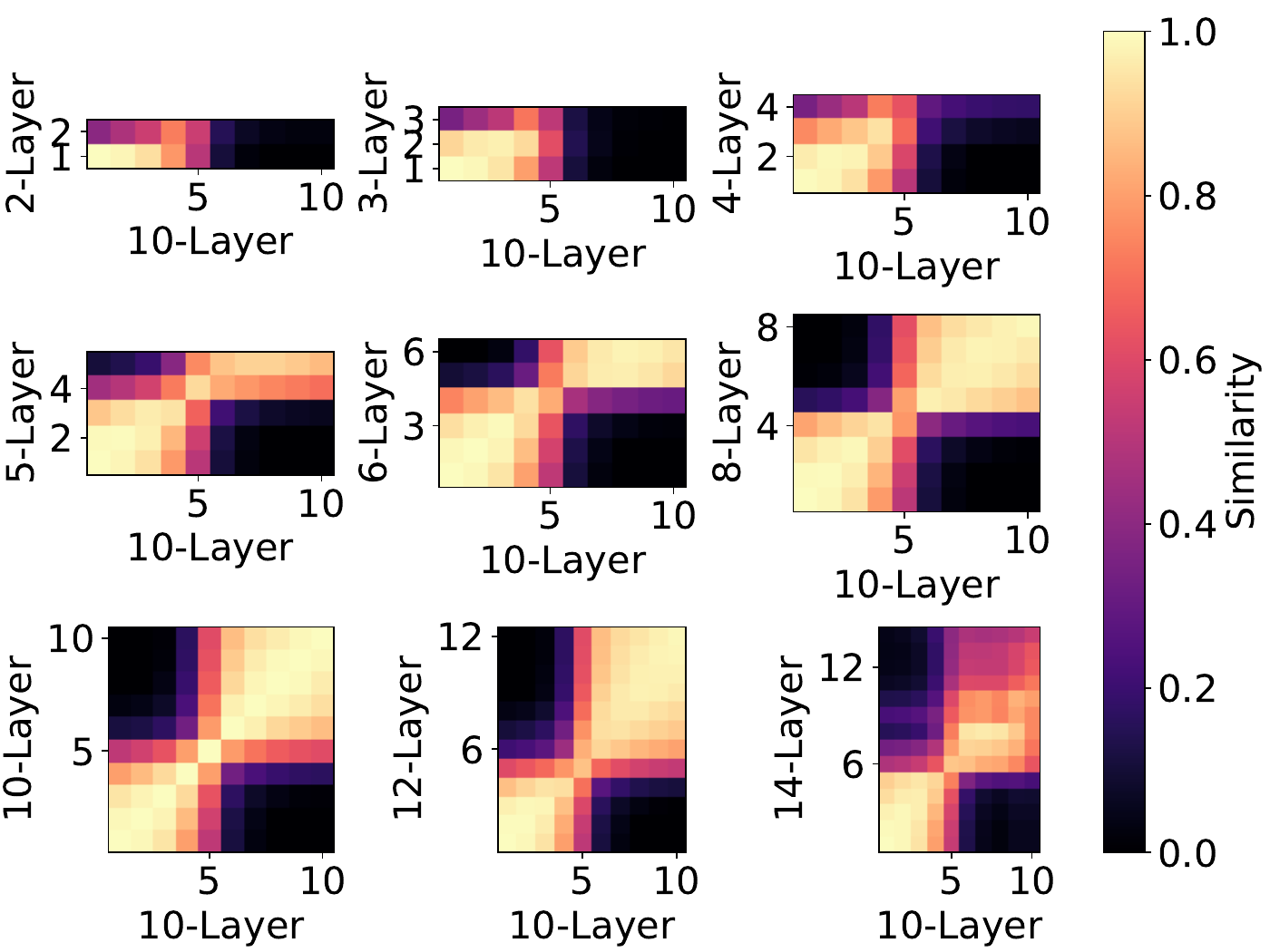}\label{fig:cka-10cross}}
    \subfloat[5-to-all]{\includegraphics[width=.5\linewidth]{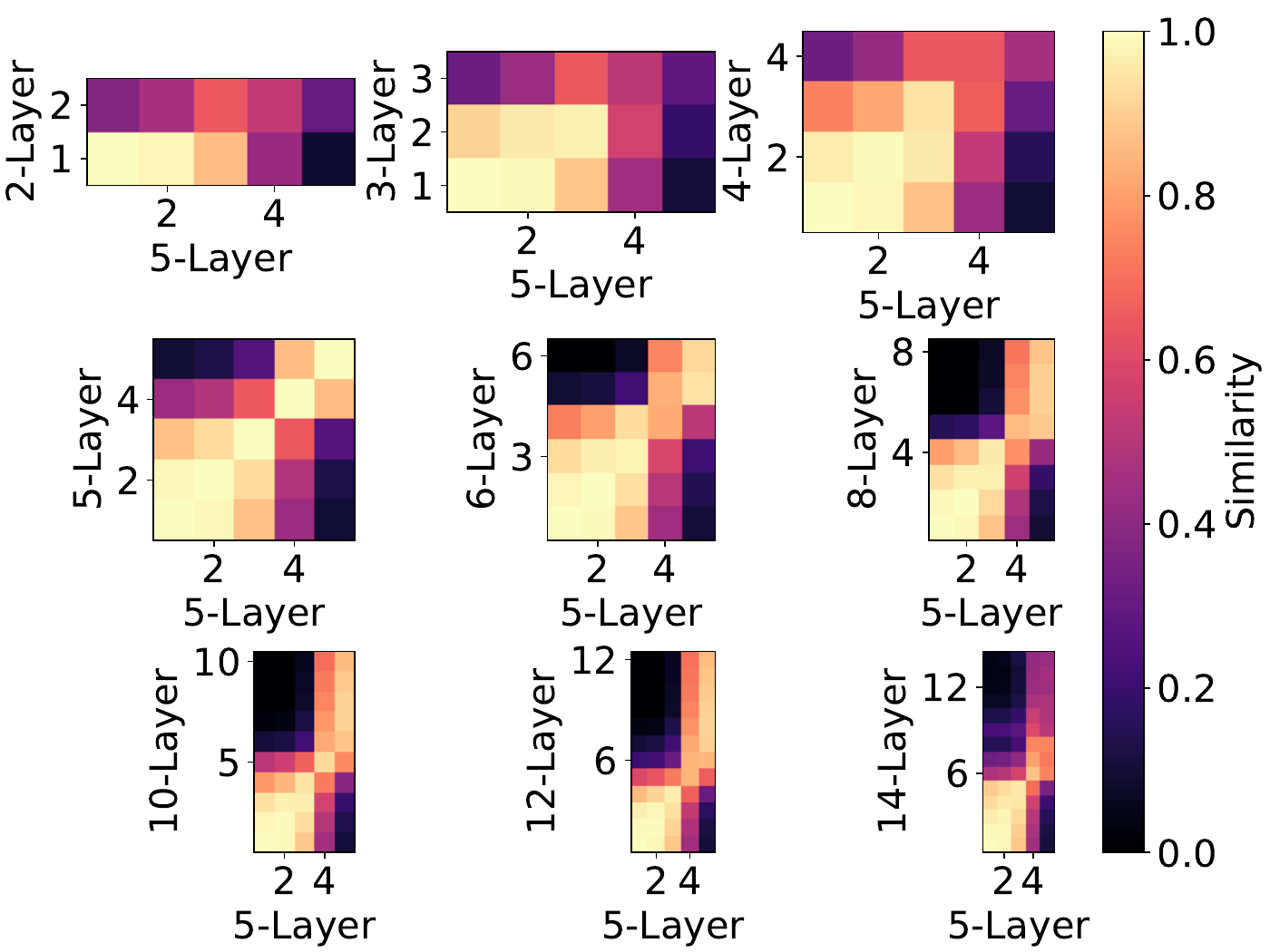}\label{fig:cka-5cross}}
    \caption{CKA 10-to-all (\ref{fig:cka-10cross}) and 5-to-all (\ref{fig:cka-5cross}) cross-similarity comparison for 3DCNN-based architectures on the PURE dataset.}
    \label{fig:cka-3dcnn-cross}
\end{figure*}

We investigate the representations of the data by the networks using CKA as described in Section~\ref{sec:cka_method}. Figure \ref{fig:cka-self:3DCNN} shows a CKA self-comparison of 3DCNN-based models trained on the PURE dataset. This and all other CKA plots in this paper depict similarities that are averaged across the 5 folds, and we did not observe any significant differences between folds that this averaging would mask. We observe that the deeper 3DCNN variants tend to have two or three blocks of highly similar layers. This suggests that there are a limited number of distinct sections of the network performing discrete tasks, each of which merely gains new layers in a piecemeal fashion as additional layers are added.

Figure~\ref{fig:cka-10cross} confirms that these distinct model sections perform the same function even across architectures of differing depths. We begin the cross-architecture comparison with the 10-layer model because that is the depth of the published PhysNet-3DCNN architecture. In this 10-layer model, we observe three distinct regions: layers 1-4, layer 5, and layers 6-10. Interestingly, it does not appear that these regions are present at every depth, but rather that they are added only in sufficiently deep models: while the region composed of layers 1-4 has strongly similar counterparts in all compared models, the region composed of layers 6-10 does not have highly similar regions in models of 4 layers or shallower, while layer 5 has only weakly similar counterparts in shallower models. This may indicate that models of depths 2-4 are not sufficiently parameterized to gain the functionality of the latter parts of the 10-layer model.

Indeed, when comparing the 5-layer model to the other architectures in Figure~\ref{fig:cka-5cross}, we observe that the model region in layer 4 has only weak counterparts in less-deep architectures, while layer 5 does not appear to be represented at all. In contrast, when comparing the 5-layer model to deeper architectures, it appears to strongly capture the latter regions of these architectures until reaching the 14-layer model. Due to these observations, we suspect that the 5-layer model sufficiently captures the representations present in deeper models, yet without as many redundancies.

\begin{figure}
    \centering
    \includegraphics[width=\linewidth]{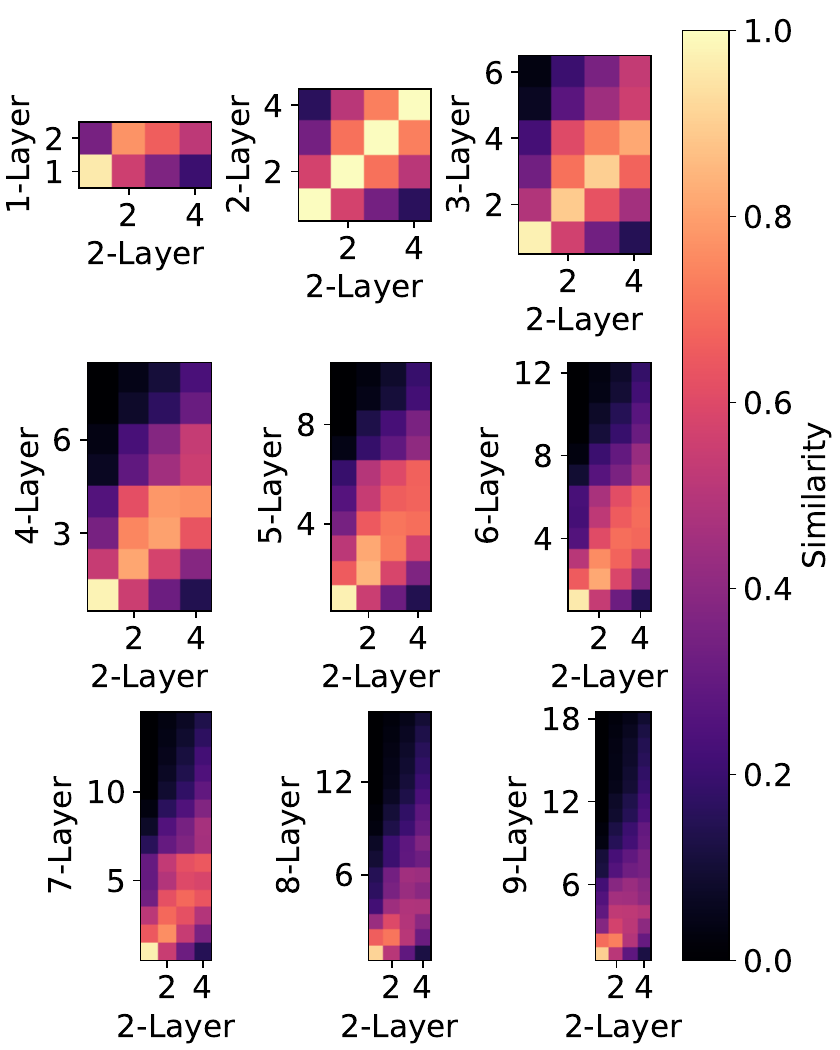}
    \caption{CKA 2-to-all cross-similarity comparison for TS-CAN based architectures on the PURE dataset.}
    \label{fig:tscan-2cross}
\end{figure}

\begin{figure*}
    \centering
    \subfloat[PURE (3DCNN)]{\includegraphics[width=.2\linewidth]{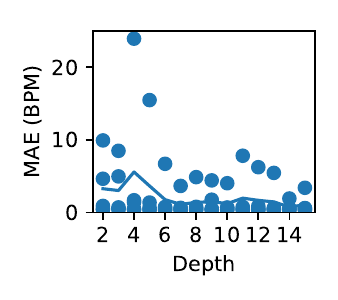}\label{fig:empirical-PURE}}
    \subfloat[UBFC-rPPG (3DCNN)]{\includegraphics[width=.2\linewidth]{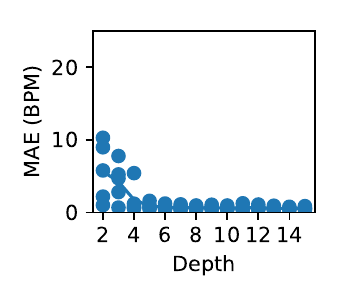}\label{fig:empirical-UBFC}}
    \subfloat[MSPM (3DCNN)]{\includegraphics[width=.2\linewidth]{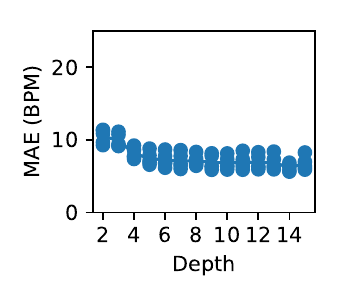}\label{fig:empirical-MSPM}}
    \subfloat[DDPM (3DCNN)]{\includegraphics[width=.2\linewidth]{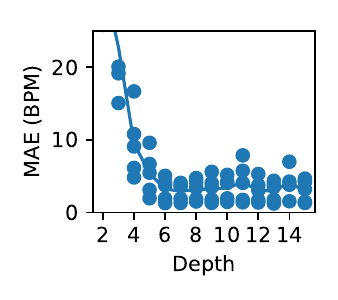}\label{fig:empirical-DDPM}}
    \subfloat[BP4D+ (3DCNN)]{\includegraphics[width=.2\linewidth]{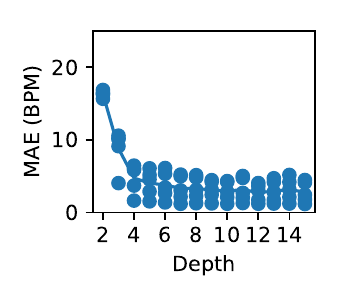}\label{fig:empirical-BP4D}}\\
    \subfloat[PURE (TS-CAN)]{\includegraphics[width=.2\linewidth]{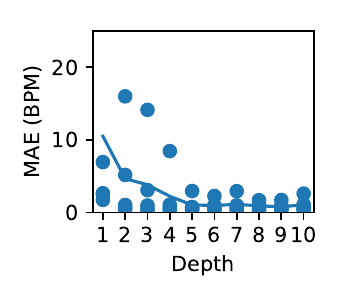}\label{fig:empirical-tscan-PURE}}
    \subfloat[UBFC-rPPG (TS-CAN)]{\includegraphics[width=.2\linewidth]{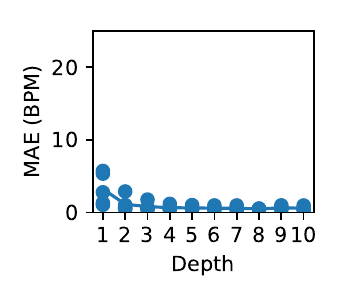}\label{fig:empirical-tscan-UBFC}}
    \subfloat[MSPM (TS-CAN)]{\includegraphics[width=.2\linewidth]{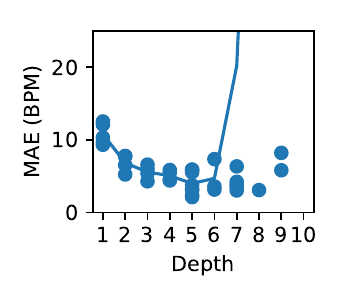}\label{fig:empirical-tscan-MSPM}}
    \subfloat[DDPM (TS-CAN)]{\includegraphics[width=.2\linewidth]{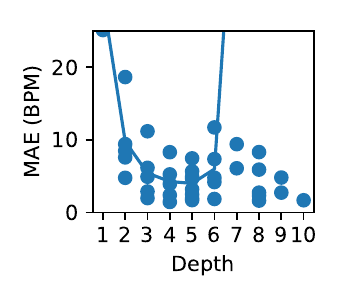}\label{fig:empirical-tscan-DDPM}}
    \subfloat[BP4D+ (TS-CAN)]{\includegraphics[width=.2\linewidth]{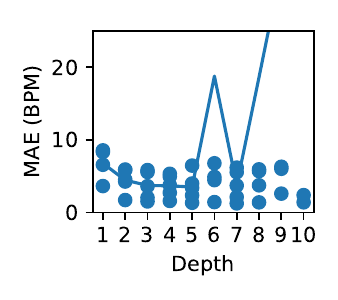}\label{fig:empirical-tscan-BP4D}}
    \caption{Empirical Results for architectures based on 3DCNN (Figures \ref{fig:empirical-PURE}-\ref{fig:empirical-BP4D}) and TS-CAN (Figures \ref{fig:empirical-tscan-PURE}-\ref{fig:empirical-tscan-BP4D}). For visualization purposes the y axis was constrained to errors under 25 BPM, which resulted in truncated results for Depths 2 and 3 for~\ref{fig:empirical-DDPM} and depth 1 in~\ref{fig:empirical-tscan-PURE} and~\ref{fig:empirical-tscan-DDPM}. Depths 7-10 in~\ref{fig:empirical-tscan-MSPM} and~\ref{fig:empirical-tscan-DDPM}, and Depths 6 and 8-10 in \ref{fig:empirical-tscan-BP4D} signal training divergence at those depths (see text for comments).}
    \label{fig:empirical}
\end{figure*}

We continue our analysis by investigating the TS-CAN architecture in Figure~\ref{fig:cka-self:TSCAN}. Because we are focused exclusively on the temporal rPPG problem, whereas TS-CAN was developed to handle multi-task learning of both appearance and temporal features, we constrain our analysis to the temporal branch, which contains two Conv2d layers for every replicated TS-CAN layer in depth. Unlike the 3DCNN architecture, TS-CAN exhibits a strong CKA diagonal with only a subtle block structure visible in Figure~\ref{fig:cka-self:TSCAN}. This may indicate that deeper variants of TS-CAN will learn more detailed representations of the data.

We continue our analysis comparing the 2-metalayer TS-CAN architecture (i.e., the published architecture) to other depths in Figure~\ref{fig:tscan-2cross}. We observe that the four Conv2d layers present exhibit the strongest similarity to the first four Conv2d layers in each deeper architecture, with low similarity to the deepest layers in the deeper achitectures. This corroborates with the self-similarity observation, that deeper TS-CAN variants appear to learn representations that are not learned by the 2-metalayer architecture.

\subsection{Empirical Error-based Analysis}

We perform an empirical study to test our findings from Section~\ref{sec:cka-results}. Figures \ref{fig:empirical-PURE}-\ref{fig:empirical-BP4D} show the Mean Absolute Error (MAE) for 3DCNN-based networks of depths 2-15 on the investigated datasets. We observe that shallower models tend to exhibit reduced accuracy and greater variation in accuracy than deeper models, corroborating the CKA diagnostic that these shallower models do not have the same level of functionality as the deeper variants. Furthermore, no significant gains appear to be made for models deeper than four layers for BP4D+ (Figure \ref{fig:empirical-BP4D}), five layers for UBFC-rPPG (Figure~\ref{fig:empirical-UBFC}), or six layers for PURE (Figure \ref{fig:empirical-PURE}) and DDPM (Figure~\ref{fig:empirical-DDPM}), corroborating the findings from the CKA analysis. There does appear to be an insignificant improvement in MSPM at 14 layers (Figure~\ref{fig:empirical-MSPM}), which could be due to the third block of layers suggested by CKA that emerges in the latter layers of 14-layer models.

We additionally test our CKA findings on TS-CAN architectures with rPPG experiments, with results documented in Figures \ref{fig:empirical-tscan-PURE}-\ref{fig:empirical-tscan-BP4D}. Our CKA findings suggested that a TS-CAN architecture of only 2 metalayers may be under-provisioned, with deeper variants learning representations that are not present in shallower models, resulting in generally lower empirical errors up to a depth of about 5 metalayers. Beyond this depth, we found that models had difficulty converging on MSPM, DDPM, and BP4D+. These datasets are larger than PURE and UBFC-rPPG by over an order of magnitude (1480, 776, and 1285 minutes for MSPM, DDPM, and BP4D+ respectively, and 60 and 70 minutes for PURE and UBFC respectively). They are also more complex, exhibiting conversation, unconstrained head movement, and activities designed to produce large fluctuations in heart rate. Meanwhile, the dual-branched TS-CAN architecture was designed to contain only two metalayers, yet we have abused its design by replicating its highly engineered structure to several times its original depth. Though we attempted extending training from 40 to 80 epochs as was done with deeper architectures based on 3DCNN, this did not result in model convergence. We believe it likely that reliably training deeper versions of TS-CAN on these more complicated datasets may require adding skip connections, adjusting dropout probabilities, learning rate scheduling, or other techniques for model convergence of deep networks.

The results for PURE in Figures~\ref{fig:empirical-PURE} and~\ref{fig:empirical-tscan-PURE} exhibit severe errors in the split containing a subject with a low heart rate and strong dichrotic notch. This contributed the highest errors for this dataset for both networks of every depth other than 3DCNN at depths 2 and 3, for which it was the 2nd highest.
%The challenge posed by this single subject in PURE has been noted by other authors~[\todo{CITE}]
\section{Conclusions}

We performed an in-depth CKA analysis on two rPPG model architectures (3DCNN and TS-CAN) over a range of architecture depths. We showed that CKA is useful for understanding model representations, both in terms of how layers are similar or unique within a model as well as across architectures, thereby informing architecture selection.

Our results, both utilizing CKA and by empirical errors across five rPPG datasets, suggest that the investigated architectures may be refined in terms of depth: The published 3DCNN depth of 10 layers appears to be deeper than necessary, with only 5 layers maintaining highly similar CKA model representations and 6 layers achieving comparable empirical performance. Similarly, the published TS-CAN depth of 2 metalayers appears to be shallower than optimal, with deeper models learning new representations not present in the published model, and empirical results showing an improved performance up to 5 metalayers.

We believe that future work utilizing CKA for architecture insights should extend in at least two dimensions. First, CKA can be used to inform more than just model depth. For example, we observe that the 3DCNN architecture exhibits a block structure at most depths, where the primary functions of the network are constrained to only a few large blocks. This is indicative that architectural adjustments other than network depth may prove fruitful in reducing model complexity without reducing performance, e.g., adjusting pooling layers, kernel sizes, or channels. Secondly, while our experiments focused on comparisons across architectures while using different datasets to validate our results, we believe that valuable insight could be gained by comparing models trained on different datasets using CKA with regards to dataset similarity under the investigated architecture and training regime.

{\small
\bibliographystyle{ieeenat_fullname}
\bibliography{refs}

\begin{thebibliography}{28}
\providecommand{\natexlab}[1]{#1}
\providecommand{\url}[1]{\texttt{#1}}
\expandafter\ifx\csname urlstyle\endcsname\relax
  \providecommand{\doi}[1]{doi: #1}\else
  \providecommand{\doi}{doi: \begingroup \urlstyle{rm}\Url}\fi

\bibitem[Bobbia et~al.(2019)Bobbia, Macwan, Benezeth, Mansouri, and
  Dubois]{Bobbia2019}
Serge Bobbia, Richard Macwan, Yannick Benezeth, Alamin Mansouri, and Julien
  Dubois.
\newblock {Unsupervised skin tissue segmentation for remote
  photoplethysmography}.
\newblock \emph{Pattern Recognition Letters}, 124:\penalty0 82--90, 2019.

\bibitem[Chen and McDuff(2018)]{Chen2018}
Weixuan Chen and Daniel McDuff.
\newblock {DeepPhys}: Video-based physiological measurement using convolutional
  attention networks.
\newblock In \emph{European Conference on Computer Vision (ECCV)}, pages
  356--373, 2018.

\bibitem[Cui et~al.(2022)Cui, Kumar, Marttinen, and Kaski]{cui2022deconfounded}
Tianyu Cui, Yogesh Kumar, Pekka Marttinen, and Samuel Kaski.
\newblock Deconfounded representation similarity for comparison of neural
  networks.
\newblock \emph{Advances in Neural Information Processing Systems},
  35:\penalty0 19138--19151, 2022.

\bibitem[{de Haan} and {Jeanne}(2013)]{DeHaan2013}
G. {de Haan} and V. {Jeanne}.
\newblock Robust pulse rate from chrominance-based rppg.
\newblock \emph{IEEE Trans. on Biom. Eng.}, 60\penalty0 (10):\penalty0
  2878--2886, 2013.

\bibitem[Kornblith et~al.(2019)Kornblith, Norouzi, Lee, and
  Hinton]{kornblith2019similarity}
Simon Kornblith, Mohammad Norouzi, Honglak Lee, and Geoffrey Hinton.
\newblock Similarity of neural network representations revisited.
\newblock In \emph{International conference on machine learning}, pages
  3519--3529. PMLR, 2019.

\bibitem[Kriegeskorte et~al.(2008)Kriegeskorte, Mur, and
  Bandettini]{kriegeskorte2008representational}
Nikolaus Kriegeskorte, Marieke Mur, and Peter~A Bandettini.
\newblock Representational similarity analysis-connecting the branches of
  systems neuroscience.
\newblock \emph{Frontiers in systems neuroscience}, 2:\penalty0 249, 2008.

\bibitem[Laakso and Cottrell(2000)]{laakso2000content}
Aarre Laakso and Garrison Cottrell.
\newblock Content and cluster analysis: assessing representational similarity
  in neural systems.
\newblock \emph{Philosophical psychology}, 13\penalty0 (1):\penalty0 47--76,
  2000.

\bibitem[Lee et~al.(2020)Lee, Chen, and Lee]{Lee_ECCV_2020}
Eugene Lee, Evan Chen, and Chen-Yi Lee.
\newblock Meta-rppg: Remote heart rate estimation using a transductive
  meta-learner.
\newblock In \emph{European Conference on Computer Vision (ECCV)}, 2020.

\bibitem[Liu et~al.(2020)Liu, Fromm, Patel, and McDuff]{liu2020multi}
Xin Liu, Josh Fromm, Shwetak Patel, and Daniel McDuff.
\newblock Multi-task temporal shift attention networks for on-device
  contactless vitals measurement.
\newblock \emph{Advances in Neural Information Processing Systems},
  33:\penalty0 19400--19411, 2020.

\bibitem[Liu et~al.(2022)Liu, Narayanswamy, Paruchuri, Zhang, Tang, Zhang,
  Wang, Sengupta, Patel, and McDuff]{liu2022rppg}
Xin Liu, Girish Narayanswamy, Akshay Paruchuri, Xiaoyu Zhang, Jiankai Tang,
  Yuzhe Zhang, Yuntao Wang, Soumyadip Sengupta, Shwetak Patel, and Daniel
  McDuff.
\newblock rppg-toolbox: Deep remote ppg toolbox.
\newblock \emph{arXiv preprint arXiv:2210.00716}, 2022.

\bibitem[Lu et~al.(2021)Lu, Han, and Zhou]{lu2021dual}
Hao Lu, Hu Han, and S~Kevin Zhou.
\newblock Dual-gan: Joint bvp and noise modeling for remote physiological
  measurement.
\newblock In \emph{Proceedings of the IEEE/CVF conference on computer vision
  and pattern recognition}, pages 12404--12413, 2021.

\bibitem[Morcos et~al.(2018)Morcos, Raghu, and Bengio]{morcos2018insights}
Ari Morcos, Maithra Raghu, and Samy Bengio.
\newblock Insights on representational similarity in neural networks with
  canonical correlation.
\newblock \emph{Advances in neural information processing systems}, 31, 2018.

\bibitem[Niu et~al.(2023)Niu, Speth, Vance, Sporrer, Czajka, and
  Flynn]{niu2023full}
Lu Niu, Jeremy Speth, Nathan Vance, Benjamin Sporrer, Adam Czajka, and Patrick
  Flynn.
\newblock Full-body cardiovascular sensing with remote photoplethysmography.
\newblock In \emph{Proceedings of the IEEE/CVF Conference on Computer Vision
  and Pattern Recognition}, pages 5993--6003, 2023.

\bibitem[Niu et~al.(2020)Niu, Shan, Han, and Chen]{Niu2020}
Xuesong Niu, Shiguang Shan, Hu Han, and Xilin Chen.
\newblock {RhythmNet}: End-to-end heart rate estimation from face via
  spatial-temporal representation.
\newblock \emph{IEEE Transactions on Image Processing}, 29:\penalty0
  2409--2423, 2020.

\bibitem[Poh et~al.(2010)Poh, McDuff, and Picard]{poh2010non}
Ming-Zher Poh, Daniel~J McDuff, and Rosalind~W Picard.
\newblock Non-contact, automated cardiac pulse measurements using video imaging
  and blind source separation.
\newblock \emph{Optics express}, 18\penalty0 (10):\penalty0 10762--10774, 2010.

\bibitem[Raghu et~al.(2017)Raghu, Gilmer, Yosinski, and
  Sohl-Dickstein]{raghu2017svcca}
Maithra Raghu, Justin Gilmer, Jason Yosinski, and Jascha Sohl-Dickstein.
\newblock Svcca: Singular vector canonical correlation analysis for deep
  learning dynamics and interpretability.
\newblock \emph{Advances in neural information processing systems}, 30, 2017.

\bibitem[Speth et~al.(2021)Speth, Vance, Flynn, Bowyer, and
  Czajka]{speth2021unifying}
Jeremy Speth, Nathan Vance, Patrick Flynn, Kevin Bowyer, and Adam Czajka.
\newblock Unifying frame rate and temporal dilations for improved remote pulse
  detection.
\newblock \emph{Computer Vision and Image Understanding}, 210:\penalty0 103246,
  2021.

\bibitem[Stricker et~al.(2014)Stricker, Muller, and Gross]{Stricker2014}
Ronny Stricker, Steffen Muller, and Horst~Michael Gross.
\newblock {Non-contact video-based pulse rate measurement on a mobile service
  robot}.
\newblock \emph{IEEE International Symposium on Robot and Human Interactive
  Communication}, pages 1056--1062, 2014.

\bibitem[Takano and Ohta(2007)]{takano2007heart}
Chihiro Takano and Yuji Ohta.
\newblock Heart rate measurement based on a time-lapse image.
\newblock \emph{Medical engineering \& physics}, 29\penalty0 (8):\penalty0
  853--857, 2007.

\bibitem[Tsou et~al.(2020)Tsou, Lee, Hsu, and Chang]{tsou2020siamese}
Yun-Yun Tsou, Yi-An Lee, Chiou-Ting Hsu, and Shang-Hung Chang.
\newblock Siamese-rppg network: Remote photoplethysmography signal estimation
  from face videos.
\newblock In \emph{Proceedings of the 35th annual ACM symposium on applied
  computing}, pages 2066--2073, 2020.

\bibitem[Vance et~al.(2022)Vance, Speth, Khan, Czajka, Bowyer, Wright, and
  Flynn]{Vance2022}
Nathan Vance, Jeremy Speth, Siamul Khan, Adam Czajka, Kevin~W. Bowyer, Diane
  Wright, and Patrick Flynn.
\newblock Deception detection and remote physiological monitoring: A dataset
  and baseline experimental results.
\newblock \emph{IEEE Transactions on Biometrics, Behavior, and Identity Science
  (TBIOM)}, pages 1--1, 2022.

\bibitem[Vance et~al.(2023)Vance, Speth, Sporrer, and
  Flynn]{vance2023promoting}
Nathan Vance, Jeremy Speth, Benjamin Sporrer, and Patrick Flynn.
\newblock Promoting generalization in cross-dataset remote
  photoplethysmography.
\newblock In \emph{Proceedings of the IEEE/CVF Conference on Computer Vision
  and Pattern Recognition}, pages 5984--5992, 2023.

\bibitem[Verkruysse et~al.(2008)Verkruysse, Svaasand, and
  Nelson]{verkruysse2008remote}
Wim Verkruysse, Lars~O Svaasand, and J~Stuart Nelson.
\newblock Remote plethysmographic imaging using ambient light.
\newblock \emph{Optics express}, 16\penalty0 (26):\penalty0 21434--21445, 2008.

\bibitem[{Wang} et~al.(2017){Wang}, {den Brinker}, {Stuijk}, and {de
  Haan}]{Wang2017}
W. {Wang}, A.~C. {den Brinker}, S. {Stuijk}, and G. {de Haan}.
\newblock Algorithmic principles of remote ppg.
\newblock \emph{IEEE Trans. on Biom. Eng.}, 64\penalty0 (7):\penalty0
  1479--1491, 2017.

\bibitem[Yu et~al.(2019)Yu, Li, and Zhao]{yu2019remote}
Zitong Yu, Xiaobai Li, and Guoying Zhao.
\newblock Remote photoplethysmograph signal measurement from facial videos
  using spatio-temporal networks.
\newblock \emph{arXiv preprint arXiv:1905.02419}, 2019.

\bibitem[Yu et~al.(2022)Yu, Shen, Shi, Zhao, Torr, and Zhao]{Yu_2022_CVPR}
Zitong Yu, Yuming Shen, Jingang Shi, Hengshuang Zhao, Philip~H.S. Torr, and
  Guoying Zhao.
\newblock Physformer: Facial video-based physiological measurement with
  temporal difference transformer.
\newblock In \emph{Proceedings of the IEEE/CVF Conference on Computer Vision
  and Pattern Recognition (CVPR)}, pages 4186--4196, 2022.

\bibitem[Zhang et~al.(2016)Zhang, Girard, Wu, Zhang, Liu, Ciftci, Canavan,
  Reale, Horowitz, Yang, et~al.]{zhang2016multimodal}
Zheng Zhang, Jeff~M Girard, Yue Wu, Xing Zhang, Peng Liu, Umur Ciftci, Shaun
  Canavan, Michael Reale, Andy Horowitz, Huiyuan Yang, et~al.
\newblock Multimodal spontaneous emotion corpus for human behavior analysis.
\newblock In \emph{Proceedings of the IEEE conference on computer vision and
  pattern recognition}, pages 3438--3446, 2016.

\bibitem[Zhao et~al.(2021)Zhao, Zou, Yang, Lu, Belkacem, and Chen]{Zhao_2021}
Yu Zhao, Bochao Zou, Fan Yang, Lin Lu, Abdelkader~Nasreddine Belkacem, and Chao
  Chen.
\newblock Video-based physiological measurement using 3d central difference
  convolution attention network.
\newblock In \emph{2021 IEEE International Joint Conference on Biometrics
  (IJCB)}, pages 1--6, 2021.

\end{thebibliography}
}

\end{document}